\documentclass[letterpaper]{article}
\usepackage{spconf,amsmath,amssymb,graphicx}
\usepackage[ruled,vlined]{algorithm2e}
\usepackage{booktabs}

\usepackage[
    backend=biber,
    style=ieee,
  ]{biblatex}

\newcommand{\mypar}[1]{{\bf #1.}}

\title{Fast single-core K-Nearest Neighbor Graph computation}
\name{Dan Kluser, Jonas Bokstaller, Samuel Rutz, Tobias Buner}
\address{Department of Computer Science\\ ETH Zurich, Switzerland}

\begin{document}
%\ninept
%
\maketitle

\begin{abstract}
%Fast and reliable K-Nearest Neighbor graph algorithms are more important than ever because of the rise of machine learning and data mining. To calculate the KNN of a given graph, a deterministic heuristic needs at least $O(n^2)$ because it is basically a matrix-matrix-multiplication.\\
Fast and reliable K-Nearest Neighbor Graph algorithms are more important than ever due to their widespread use in many data processing techniques. This paper presents a runtime optimized C implementation of the heuristic "NN-Descent" algorithm by Wei Dong et al. \cite{nndescent} for the l2-distance metric.
Various implementation optimizations are explained which improve performance for low-dimensional as well as high dimensional datasets.\\
Optimizations to speed up the selection of which datapoint pairs to evaluate the distance for are primarily impactful for low-dimensional datasets. A heuristic which exploits the iterative nature of NN-Descent to reorder data in memory is presented which enables better use of locality and thereby improves the runtime. The restriction to the l2-distance metric allows for the use of blocked distance evaluations which significantly increase performance for high dimensional datasets.\\
In combination the optimizations yield an implementation which significantly outperforms a widely used implementation of NN-Descent on all considered datasets. For instance, the runtime on the popular MNIST handwritten digits dataset is halved.

% To overcome this runtime bottleneck we used a novel approach called "NNDescent" which was introduced by Wei Dong et al. \cite{nndescent} and uses a randomized iterative heuristic. The heuristic of this approach is that neighbors of neighbors are more likely to also be neighbors. \\
% As baseline we used Skomski's C implementation \cite{Eskomski19} of the previously mentioned paper. To improve the random initialization, we used some concepts of the Python implementation called "PyNNDescent" \cite{pynndescent} for a 16x speedup, by further improving these concepts we gained another 1.12x speedup.
% To improve the actual computation loop, which updates the KNN of every point, we utilized intrinsics for the L2-norm, blocking for pairwise distances in 5x5 blocks and memory alignment. By assuming clustered input-data we were able to get most near neighbors also near in memory, this helps us overcome the main problem, that neighbor-of-neighbor access pattern takes us anywhere in memory.\\
% With these improvements in place we were able to improve the runtime significantly and also beat "PyNNDescent" \cite{pynndescent} in synthetic and real data load.
\end{abstract}

\section{Introduction}\label{sec:introduction}
K-nearest neighbor graphs (K-NNG), which contain the identities of the closest datapoints for each datum, are fundamental to many data science and machine learning techniques. The rapid growth of attainable data has therefore increased the need for practical K-NNG algorithms and efficient implementations thereof.\\
This paper presents an optimized implementation of the NN-Descent algorithm by Wei Dong et al. \cite{nndescent}. NN-Descent is an iterative, randomized heuristic which improves a random guess of the K-NNG.\\
A particularly popular implementation of NN-Descent is PyNNDescent \cite{pynndescent} which is implemented in Python and uses the Numba JIT copiler \cite{noauthor_numba_nodate} to achieve sufficient runtime performance. Particular strengths of PyNNDescent are its ease of use and support of custom distance metrics. PyNNDescent is used in the sci-kit learn compatible implementation of UMAP \cite{mcinnes_lmcinnesumap_2020} \cite{umap}. UMAP is currently gaining popularity as an alternative dimensionality reduction technique to t-SNE due to favorable embedding properties and because it is faster than current implementations of t-SNE \cite{becht_dimensionality_2019}.\\
The use of NN-Descent in practice makes performance optimized implementations especially desirable.\\
A major challenge in optimizing the performance of NN-Descent implementations is the irregular memory access pattern stemming from unordered input data together with the neighbor-of-neighbor heuristic explained in Section \ref{sec:nn_descent_algo}.\\
This paper presents a single-core implementation which is limited to the l2-distance metric at the benefit of a vast runtime reduction when compared to PyNNDescent.\\
Optimization techniques are discussed which performance for both low or high-dimensional input respectively. In particular a novel heuristic is introduced which increases locality by improving the otherwise irregular memory access pattern.\\
The overall performance advantage over PyNNDescent motivates the use of specialized implementations of NN-Descent for commonly used distance metrics and use cases.\\

Algorithms for K-NNG computation are still an active area of research due to their enormous importance. Alternative approaches include "Goldfinger" \cite{Ruas} which is based on bit-comparisons of hashes as well as methods building upon the same ideas as NN-Descent \cite{baron}.

\section{K-Nearest Neighbor Graph and NN-Descent}\label{sec:nn_descent_algo}
In this section we present an overview of the K-Nearest Neighbor Graph (K-NNG) problem and the NN-Descent algorithm optimized in our implementation.\\

%A K-NNG contains a set of the $k \in \N$ closest vertices for each vertex in a dataset, according to a given distance function.
A K-NNG is a directed simple graph $G = (V,E)$ such that we have a directed edge $(u,v) \in E$ iff $v$ is one of the k-nearest neighbors of $u$ according to a given distance metric. Let $V$ be the set of datapoints in $\mathbb{R}^{d}$ space where $|V| = n$. The dimensionality of all datapoints is denoted by $d$. The $k$ nodes adjacent to some node $u$ in $G$, denoted as $\text{adj}_{G}(u)$, are thus the $k$ nearest neighbors according to the given metric.

%For a given datapoint $v$ we define the set $B^{*}(v)$ to contain the indices of the $k$ datapoints in $V$ for which the distance to $v$ is smallest (excluding $v$ itself).\\
One trivial way of computing the KNN-Graph is to compute all $\frac{n(n-1)}{2}$ mutual distances and selecting the best $k$ points. This is not computationally viable for most datasets of practical interest, especially in data science applications due to the $O(n^{2})$ asymptotic number of distance evaluations.\\
NN-Descent by Wei Dong et al.\cite{nndescent} is a randomized, iterative heuristic which computes an approximation of the K-NNG. The NN-Descent algorithm requires far fewer distance evaluations (empirical cost $O(n^{1.14})$ \cite{nndescent}) at the expense of the quality of the resulting KNN-Graph.\\

\begin{figure}\centering
  \includegraphics[width=0.3\textwidth]{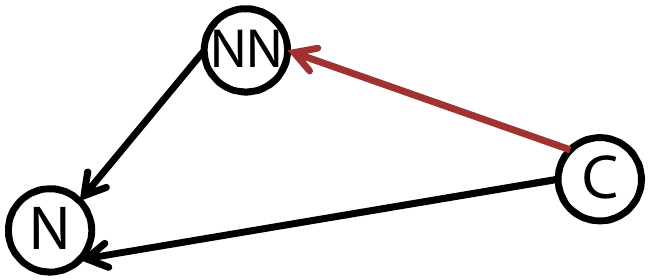}
  \caption{A neighbor of a neighbor is likely to be also a neighbor \cite{nndescent}. In the current estimate of the K-NNG, N is considered a neighbor of NN and also of C. NN-Descent evaluates the distance between NN and C updates the current approximation of the K-NNG accordingly by replacing edge $($C,N$)$ with $($C,NN$)$. \label{fig:neighbor}}
\end{figure}

The key insight used is that "a neighbor of a neighbor is also likely to be a neighbor"\cite{nndescent} (Figure \ref{fig:neighbor}).

Using this assumption allows NN-Descent to iteratively improve the KNN-Graph estimate by evaluating distances primarily for promising pairs of vertices.\\
NN-Descent begins with a random initialization of $G$, so that for every vertex the set of $k$ nearest neighbors are uniformly sampled from $V$. Multiple iterations of the NN-Descent algorithm are then executed. Every iteration improves the K-NNG estimate by performing the following two steps for every vertex in $V$:
\begin{enumerate}
    \item Select candidates: find a suitable set $S$ with vertices that are neighbors-of-neighbors
    \item Calculate and update: calculate all pairwise distances in $S$ and update $G$ if closer vertices were found.
\end{enumerate}
We will refer to the first step as \textit{Selection step} and the second one as \textit{Calculation step}. Iterations are performed until the number of changes to $G$ falls below a specified threshold. In this case the estimated K-NNG is a reasonable approximation of the true K-NNG.\\
Section \ref{subsec:selection_step} contains more details on the selection of candidates. For more details on the algorithm itself refer to \cite{nndescent}.\\
As the NN-Descent algorithm is a heuristic it is necessary to validate both the quality of the returned K-NNG as well as the computational cost. Recall is used to measure how close the K-NNG approximation is to the true K-NNG. Our implementation achieved a recall of over $99\%$ on all examined datasets. Multiple parameters could if desired be altered to change the runtime-quality trade-off.\\
To calculate the number of floating point operations, the number of distance evaluations is counted. This enables performance comparisons between different code versions which have small but varying amounts of additional floating point comparisons. Together with the dimensionality of the datapoints the number of operations can be computed. For each l2-distance evaluation $d$ subtractions, $d$ multiplications, and $d-1$ additions are performed. 

\section{Optimization of NN-Descent implementation}\label{sec:yourmethod}
Numerous possible optimizations were implemented and evaluated to yield a fast single-core implementation of NN-Descent. Some approaches aim at making the selection step highly efficient while others aim to improve the locality or optimize the mutual distance calculations. The relative impact of these optimizations differs depending on parameters of the problem, such as the dimensionality $d$ of the datapoints.\\

We started out with a C implementation\cite{Eskomski19} which adheres closely to the pseudo code found in \cite{nndescent} (NNDescentFull). We then studied the most popular implementation of NN-Descent,  PyNNDescent \cite{pynndescent}, which was introduced in Section \ref{sec:introduction}.
Although asymptotically the two implementations behave the same, the Python implementation was considerably faster than the straightforward C implementation. We identified the main differences and adopted the improvements.% into the straightforward implementation.% up to speed.

\subsection{Selection Step}\label{subsec:selection_step}
An important improvement found in PyNNDescent is the selection step, which is performed in each iteration for every vertex (see Section \ref{sec:nn_descent_algo}). These improvements were then expanded upon for an additional speed-up.

In the selection step of every iteration we need to find the neighborhood $N(u)$ for every node $u$ in the current KNN-graph approximation. The neighborhood $N(u)$ contains every node $v \in V$ for which $(u,v) \in E$, i.e. node $v$ is one of the k-nearest neighbors according to the current approximation. More interestingly, it also contains every node $w$ for which $(w,u) \in E$, so every node $w$ which has node $u$ as one of its k-nearest neighbors.
In the pseudo code as presented \cite{nndescent} such nodes $w$ are found by first inverting all the directed edges in the current KNN-graph $G$, resulting in a graph $G' = (V, E')$ for which $E' = \{ (u,v) | (v,u) \in E\}$. After the \textit{reverse} step, we then proceed by setting $N(u) = \text{adj}_{G}(u) \cup \text{adj}_{G'}(u)$ for all $u$ which we refer to as \textit{union} step. Note that while $\text{adj}_{G}(u)$ is bounded in size by k, $\text{adj}_{G'}(u)$ can contain up to $n$ elements, which requires the usage of a dynamically growing data structure.

Finally we \textit{sample} the neighborhood $N(u)$ to contain $\rho \cdot k$ elements. This reduces the number of the pairwise distance calculations of the neighborhood $N(u)$ which is quadratic in $|N(u)|$. The whole process of finding a suitably sized neighborhood for every node can be regarded as a composition of three functions: \textit{reverse, union} and \textit{sample}. In the basic implementation the intermediate results are stored in memory and three full passes over the entire K-NNG are required.

PyNNDescent improves over this by only doing one pass over the data but introducing a heap in the sampling process. Instead of building each neighborhood $N(u)$ and then sampling from it, PyNNDescent does both simultaneously in one pass over the KNN-graph. We adopted this change. For each edge $r = (u,v)$ a weight $r_e$ is drawn uniformly at random (u.a.r.) in $[0,1]$. Both $N(u)$ and $N(v)$ are implemented as heaps and we insert the node $v$ in $N(u)$ with weight $r_e$, we do the symmetric thing for $N(v)$. This corresponds to both the reverse and union step of the naïve selection implementation. Because the heaps are bounded in size, every neighborhood $N(v)$ ends up containing at most $\rho \cdot k$ elements. Selecting a subset of size $\rho \cdot k$ is equivalent to assigning a random weight u.a.r. to each element and selecting the $\rho \cdot k$ elements with the smallest weights. This gives a considerable speedup; on our synthetic dataset we observed a 16x runtime speedup (see Section \ref{subsec:exp_selection}).

%Since the heaps appeared wasteful to us and incurred a lot of cache misses we further optimized the fused selection function outlined above to get rid of the heaps.
Because the heaps incurred a lot of cache misses we further optimized the fused selection function described above in order to get rid of the heaps.
Upon every update of the KNN-graph we keep track of how large the neighborhood of every node $v$ is. Since when doing these updates we access the relevant data structures anyway, we do not incur any additional cache misses by these modifications. Knowing how large each neighborhood is allows us to simplify the sampling process: for every edge $e = (u,v)$ we insert $v$ into $N(u)$ with probability $\frac{\rho \cdot k}{|N(u)|}$. In expectation this is equivalent to the previous sampling procedure, but it works without heaps. This gives a small speedup of around 1.12x.

%Note the reduced number of memory accesses by eliminating the intermediate results necessary when applying the three functions in sequence instead of the optimized composition. 
The optimized one-pass sampling step reduced the number of memory accesses by eliminating the intermediate results of the sequentially applied functions (\textit{reverse, union} and \textit{sample}).
%Furthermore we avoided the difficulty of controlling (in terms of size of adjacency lists) the reverse graph $G'$ and reduce the number of passes over the graph down to a single pass. 
Furthermore, we avoided the difficulty of controlling the (previously unbounded) size of the reverse graph $G'$.
This is important since for many relevant input sizes the KNN-graph does not fit into the caches found in commodity hardware. Multiple passes over the KNN-graph then lead to many cache misses.

% \subsection{Impact of dimensionality}\label{subsec:impactofdimensionality}
% Optimizing the selection step highlighted that a lot of the runtime for low dimensions $d$ arises from these computations as opposed to actual distance evaluations. As the computational cost of the selection step is independent of the data dimensionality $d$, the cost of the distance evaluations outweighs this for higher dimensional data sets. This shift is visible in the roofline plot \ref{fig:rooflineplot}, which is further elaborated in Section \ref{subsec:exp_roofline}.

% Using the available information we can calculate the operational intensity for different dimensions and therefore integrate them into our roofline model. For higher dimension the computation becomes compute bound instead of memory bound. So changing the focus also changes the bottleneck and therefore other improvements are more important. For high dimensions the calculation step is where we spend most time. Those improvements can be found in section \ref{subsec:highdimension} but what is possible when working with low dimension?

\subsection{Greedy Reordering Heuristic}\label{subsec:greedy}
Datapoints which are close in the dataspace are frequently accessed together but the underlying data is not usually located closely together. This leads to a difficulty in exploiting spatial locality. Our roofline model analysis (cf. Section \ref{subsec:exp_roofline}) indicates that our implementation is memory-bound for low-dimensional inputs. The major difficulty of exploiting locality in spite of the non-uniform memory access pattern is the primary issue to be solved. Introducing an assumption on the data space distribution of the input data allows the development of a heuristic approach to tackle this problem.

%Our assumption is based on the fact that the data is clustered and every nodes $k$ nearest neighbors are within the same cluster. 

Without any assumptions about the input distribution, our access pattern is irregular and we cannot improve locality. This is due to the tight relationship of temporal locality of two nodes and their distance in data space. We proceeded by assuming our input is clustered, meaning for every node all its $k$ nearest neighbors are within the same cluster (\textit{clustered assumption}). This assumption will allow us to partly recover those clusters from an early approximation of the K-NNG. After reordering memory such that the clusters are close together, we proceed with the remaining iterations of NN-Descent. Experiments on a synthetic dataset and on real world data set are promising (consult section \ref{sec:exp} for more details).

Recall that during the selection step we iterate over all edges $e \in E$ of the current KNN-graph approximation $G = (V,E)$. For one edge $e = (u,v)$ we will access both $\text{adj}_{G}(u)$ and $\text{adj}_{G}(v)$. Those two lists are likely to be in completely different locations in memory. Since the edge $e$ is part of our current KNN-graph, $u$ and $v$ are likely close in data space according to the given metric. This is why, especially after the initial iteration when our KNN-graph approximation becomes more accurate, closeness in data-space and temporal locality in the access pattern are highly correlated. For the remainder of this section we will consider clustered inputs (\textit{clustered assumption}).

Intuitively, after the first iteration a nodes nearest neighbor is likely to be in the same cluster. Recall that we start with a randomly initialized approximation, meaning every node has $k$ u.a.r. chosen nearest neighbors. The probability that within those $k$ nodes we do not have any node from the same cluster can be bounded from above by (for $c$ clusters, and $k$ neighbors):
$$\Pr[\text{all $k$ nodes not in cluster $i$}] \leq \Big (\frac{c-1}{c} \Big )^k$$

For a wide range of practical $k$ and $c$ this probability is sufficiently small. Armed with the intuition that the nearest neighbor of every node in our approximation is likely to be within the same cluster, we may now try to exploit that to reorder our memory. As a first step we want an algorithm that:
\begin{itemize}
    \item may only use the existence of those clusters, the input is \textit{not} ordered in any way revealing information about the structure of those clusters
    \item recovers most of the clusters. Moreover, it should output a permutation $\sigma: [n] \rightarrow [n]$ which we may use in the end to permute our data in memory all at once to bring the clusters together.
    \item makes at most one pass over the KNN-graph
\end{itemize}

The above requirements inform the design of our greedy clustering heuristic. In the pseudo code (Algorithm \ref{alg:greedy}), the permutations $\sigma$ and $\sigma^{-1}$ are modeled as $n$-dimensional arrays. We initialize them with the identity function: for each $i \in [n]$ we have $\sigma(i) = i$. We proceed by looking at node $i=0$. In each iteration of the outermost loop we would like to find a good candidate for the spot $i+1$, meaning whichever node permutation $\sigma$ maps onto $i+1$, it should be close in data space to node $i$. To achieve that we now sort the adjacency list of $i$ by distance, so $a_i[j]$ contains the identifier of the $j$'th closest node. Now we check whether the spot assigned to by $a_i[j]$ by permutation $\sigma$ is smaller than our current position, if so, we assume that $a_i[j]$ already has a good spot where it is close to its (data-space) neighbors in memory space. If not we check whether $a_i[j]$ already occupies the spot we would like to have it at ($\sigma(a_i[j]) = i+1$) - then we conclude the search and break out of the inner loop.

Otherwise if $\sigma(a_i[j]) > i+1$, we would like to set $\sigma$ such that $\sigma(a_i[j]) = i+1$, such that the node $a_i[j]$ will occupy spot $i+1$. Note that this is the desired outcome, the nodes on spots $i$ and $i+1$ are close together in data space. This specific sequence of two swaps in the permutation $\sigma$ and its inverse turn out to give the desired result. By creating and updating both the permutation and its inverse at the same time, we save ourselves a costly inversions of the permutation at several steps. This way we can satisfy the second requirement of only doing one pass through the KNN-graph.

The algorithm then returns a permutation $\sigma$. We proceed by permuting all of our data in memory using that permutation. Afterwards we continue with the remaining iterations of NN-Descent using the permuted memory layout. The copying itself is done all at once using $\sigma$. 

When the clustered assumption is given, it is intuitive that our heuristic will succeed in clustering most of the data. Consult section \ref{sec:exp} for an experimental evaluation on a synthetic data set. More surprisingly we have even seen a small speedup on real world datasets where the clustered assumption does not hold.

\begin{algorithm}
\SetAlgoLined
\KwResult{Permutation $\sigma$ }
 $\sigma \gets id$\;
 $\sigma^{-1} \gets id$\;
    \For{$i\gets0$ \KwTo $n-1$ }{
        $a_i \gets \text{sorted}(\text{adj}_{G}(i))$\;
        \For{$j\gets0$ \KwTo $k-1$ }{
            \uIf{$\sigma(a_i[j]) < i+1$}{
                continue\;
            }
            \uElseIf{$\sigma(a_i[j]) = i+1$}{
                break\;
            }
            \ElseIf{$\sigma(a_i[j]) > i+1$}{
                swap in $\sigma$ entries $a_i[j]$ and $\sigma^{-1}(i+1)$ \;
                swap in $\sigma^{-1}$ entries $\sigma(a_i[j])$ and $i+1$ \;
                break\;
            }
        }
    }
 
 \caption{Greedy Clustering Heuristic}\label{alg:greedy}
\end{algorithm}

%\mypar{High Dimension}
\subsection{Compute Step}\label{subsec:highdimension}
%section preview

\begin{figure}\centering
  \includegraphics[width=0.5\textwidth]{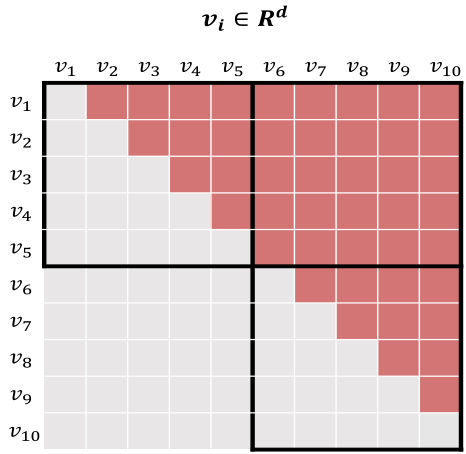}
  \caption{Illustration of blocked distance evaluations. For the neighborhood containing $v_{1}$ through $v_{10}$ the red cells represent distance evaluation. All distance evaluations within a 5 by 5 block are computed simultaneously. \label{fig:matrix}}
\end{figure}

For higher dimensional datasets each l2-distance evaluation becomes more costly while the overhead of sampling and updating data structures remains constant. In such cases optimizing the distance evaluations becomes significantly more important than the optimizations described in previous sections.\\
%This also motivated by the roofline model in section \ref{subsec:exp_roofline} where we even get compute bound for higher dimensions.
%remove sqrt
A single l2 distance evaluation for two vectors is computed by summing the component-wise distances and taking the square root. As the actual value of the l2-distance is unimportant, the square root is omitted and the implementation uses the squared l2-distance. This improvement is not significantly impactful as for high dimensional vectors the cost of computing and summing the differences is dominant.\\
%intrinsics and assumptions
We decided to limit our implementation to vector dimensions which are divisible by 8 in order to simplify the use of AVX2 SIMD intrinsics. As each AVX2 register can hold 8 single-precision floating point numbers this alleviates the need for auxiliary code to handle the last (fewer than 8) components. The real-world datasets described in Section \ref{sec:exp} happen to fulfill this requirement without modification.\\
We use a AVX2 vector of accumulators for each distance evaluation and process 8 components at a time by subtracting to compute the difference and using an \emph{fmadd} instruction to square the difference and add it to the accumulator vector (cf. tag \textit{l2intrinsics} in Section \ref{sec:exp}).\\
We noticed that the restriction of the dimensionality to divisibles by 8 allowed for an easy modification to the way the datapoints are stored in memory. This modification allocates the data neatly aligned to 256 bits at the cost of at most additional 192 bits. This change significantly improves the performance since the \emph{loadu} instructions become faster. We did not observe a speedup by replacing \emph{loadu} intrinsic instructions by the equivalent \emph{load} intrinsic (cf. tag \textit{mem-align}).
The compute step for a single node can be further improved by blocking in order to compute the mutual distances between multiple vectors simultaneously. Modifications to the basic NN-Descent algorithms sampling step lead to a neighbourhood never exceeding a fixed size (in our implementation 50 nodes). For these, all the mutual distances have to be computed which can effectively be blocked. % TODO figure
We use a blocksize of 5 by 5 vectors (not scalars) in our implementation (Figure \ref{fig:matrix}). 

For each such block we allocate an 256 bit AVX2 accumulator for every distance evaluation (either 10 or 25 in total). We then proceed with the computation of the squared l2-distance 8 features at a time for all combinations. In a block where 25 mutual distances are computed simultaneously only 10 AVX2 vectors of data are loaded per 8 dimensions. Without blocking each of theses would be loaded once for every distance evaluation. (1 vs. 25 loads per component). For high dimensional vectors especially this drastic reduction of data loads has a major impact on the performance (cf. tag \textit{blocked}).\\
Choosing a blocksize of 5x5 allows for each of the 25 accumulators to be allocated to a register which can be seen by inspecting the assembly. If the number of vectors to be compared to one-another is not divisible by 5 a more flexible but slower function is used for the remaining pairs.\\

\section{Experimental Results}\label{sec:exp}
% This section covers experimental results with respect to the previous section and the discussed topics.
This section discusses the experimental performance of the optimizations discussed in the previous section.
The main questions we aim to answer are the following:
\begin{enumerate}
    \item Does our clustering heuristic successfully cluster our data in memory?
    \item How does performance scale in terms of dataset size $n$ and dataset dimension $d$?
    %\item Do certain optimizations benefit higher dimensional data more than lower dimensional?
    \item Which optimizations are beneficial specifically in low or high dimensional settings?
    \item How does our final implementation perform to the numba based Python implementation PyNNDescent on real world data?
\end{enumerate}
 For all experiments in this section we used the squared euclidean distance and $k=20$ on single precision floats.\\

\mypar{Experimental setup} All the experiments were conducted on a computer running Ubuntu 20.04 LTS. The CPU is a Intel Core i7-9700K CPU @3.60GHz (turbo boost disabled) with the cache sizes L1: 256 KiB, L2: 2 MiB and L3: 12 MiB. The GCC version 9.3.0 was used in all experiments with the following flags: \emph{03}, \emph{ffast-math}, \emph{march=native}, \emph{flto}. The flags \emph{p} and \emph{pg} were used to add run time instrumentation to the code for specific profiling tasks.
\\

%\mypar{Datasets}\label{datasets}
Four datasets were used to evaluate the performance and recall of our modified implementation. To investigate scaling behaviours, synthetically generated datasets were used. Additionally real-world datasets including the MNIST handwritten digits \cite{mnist} were used to verify the generalization of the performance from the synthetic datasets.
%We not only used synthetic data sets, but also real world data sets like MNIST \cite{mnist}.
\begin{description}
\item[Synthetic Gaussian Dataset] The input parameter for the generation of this data set is the number of dimensions and the number of points. For the \textit{Single Gaussian Dataset} all points are drawn from one gaussian distribution centered at the origin. In the non-single variant, for each dimension a gaussian is created and centered around the canonical basis vector. For all evaluations the covariance is $2\cdot I_d$.
\item[Synthetic Clustered Dataset] A dataset designed to fulfill our clustered assumption. For every cluster we draw its points from a multivariate Gaussian. Mean and covariance are chosen such that the clustered assumption holds with high probability.
\item[MNIST Dataset \cite{mnist}] The MNIST database contains 70'000 images of handwritten digits given as 784 dimensional vectors of pixel intensity values. 
\item[Audio Dataset] The audio dataset used in the NN-Descent publication by Wei Dong et al. \cite{nndescent}. Each of the 54'387 points consists 192 features which were extracted from recordings of English sentences.
\end{description}

\subsection{Selection Step}\label{subsec:exp_selection}
Every iteration of NN-Descent consists of a selection and computation step (cf. Section \ref{sec:nn_descent_algo}). Here we evaluate the improvements detailed in Section \ref{subsec:selection_step}. As previously mentioned the PyNNDescent inspired sampling is up to 16 times faster than the naïve sampling implementation in C. Our improved sampling, which we call \textit{turbosampling}, gives a speedup on top of that of up to 1.12x. The speedup was measured in terms of runtime since the flop count varies across those three implementations and a performance plot would be misleading. We used the \textit{Synthetic Gaussian Dataset} with parameters $n=16'384$, $d=8$ for said measurements. Due to space constraints, we omit further evaluations of improvements inspired by PyNNDescent and focus on our own contributions.

\subsection{Roofline model}\label{subsec:exp_roofline}
\begin{figure}\centering
  \includegraphics[width=0.4\textwidth]{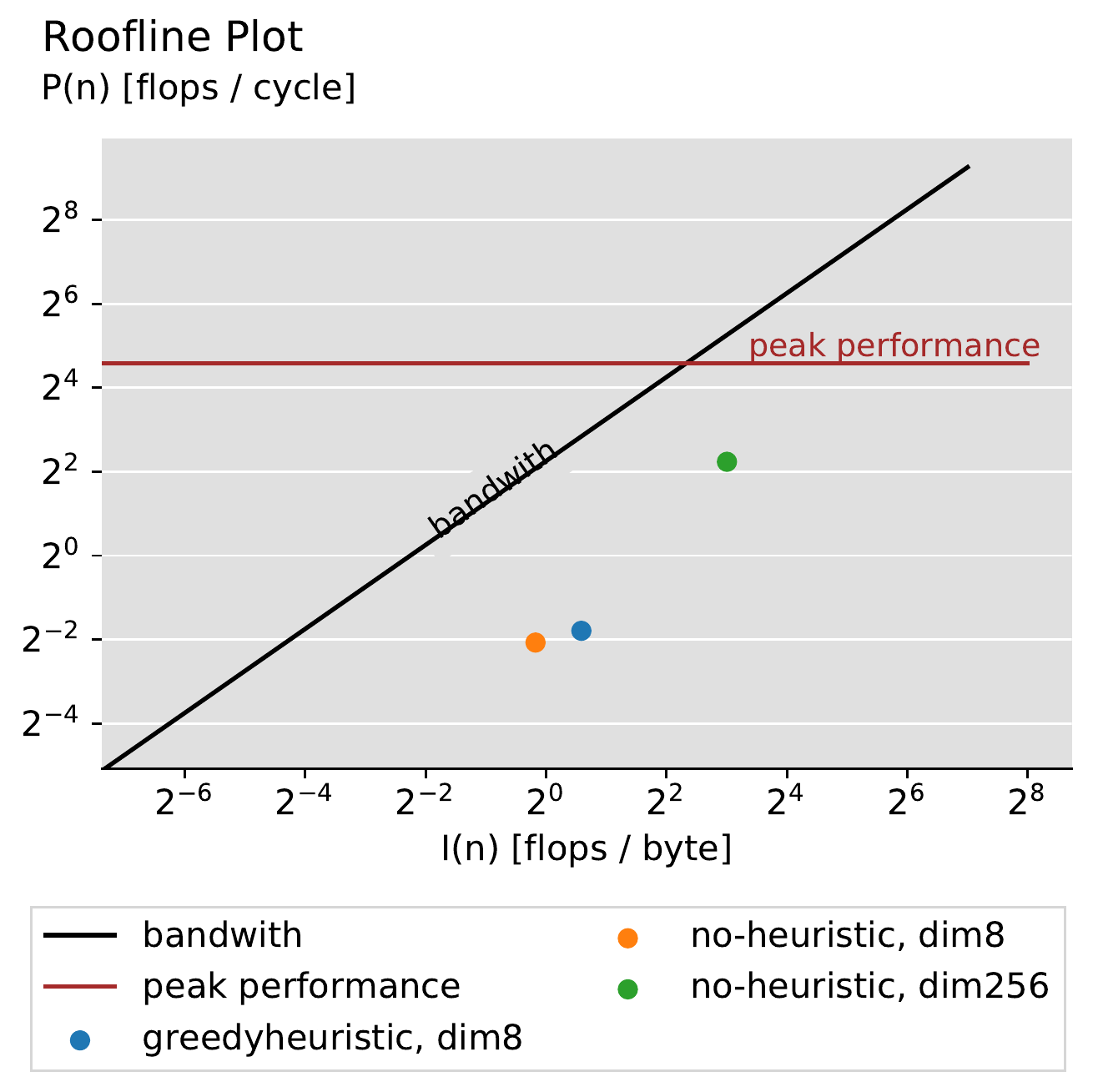}
  \caption{Roofline plot with peak performance $\pi = 24$ [flops/cycle] and bandwidth $\beta = 4.77$ [bytes/cycle]. The \textit{Synthetic Gaussian Dataset} was used with $n=131'072$ and the dimensions $8$ and $256$ respectively.}
  \label{fig:roofline}
\end{figure}
Our optimizations in Section \ref{subsec:greedy} and \ref{subsec:highdimension} are motivated by the fact that the bottleneck of our algorithm changes as we change the dimension. For low dimensional data the computation is memory bound whereas for higher dimension we actually become compute bound as shown in figure \ref{fig:roofline}. In the following the derivation of the roofline model parameters is explained.

As stated in section \ref{sec:nn_descent_algo} the work $W(n)$ is calculated from the number of distance evaluations and the dimensionality of the datapoints. 
%As stated in section \ref{sec:nn_descent_algo} we define the work $W(n)$ to be the number of floating point operations of the distance metric calculations. 
For the sake of the roofline model analysis, we additionally need to reason about the data movement $Q(n)$, the bandwidth $\beta$ as well as the peak performance $\pi$.

\mypar{Bandwidth} Using the stream benchmark tool \cite{streambenchmark}, we measured the bandwidth $\beta = 4.77$[bytes/cycle] which seems reasonable as the maximal bandwidth from the manual is rarely attainable ($41.6 GB/s$ \cite{intelcpumanual} results in $12.41$[bytes/cycle]).

\mypar{Peak performance} %Our Coffee Lake processor based on the Skylake microarchitecture achieves a peak performance of $4 [flops/cycle]$. 
%If we now introduce 8-way SIMD while working on 32-bit single precision floating points we get $32 [flops/cycle]$. 
Utilizing 8-way AVX2 SIMD instructions for single precision floating point numbers the peak performance is $32$ [flops/cycle] on the Coffee Lake processor based on the Skylake microarchitecture.
This bound does not account for the instruction mix which is about $50\%$ 8-way subtractions and $50\%$ 8-way FMA's leading to a bound of $24$ [flops/cycle] which is used as our peak performance $\pi$. 

\mypar{Data movement} The number of bytes transferred from memory to the cache far exceeds the size of the cache, which necessitates repeated loading and rewriting of data. To reason about the bytes transferred, cachegrind (Valgrind extension)\cite{cachegrind} was used. Cachegrind enables examination of the code's cache behaviour by simulating the memory behaviour in terms of first and last level (LL) caches. Using this tool we measured the LL cache data read and write misses. Table \ref{table:cachegrind} summarizes the cachegrind output on our \textit{Synthetic Clustered Dataset} with $n=131072$ and $16$ clusters.

\begin{table}\centering
\begin{tabular}{@{}ccc@{}}\toprule
& LL read misses & LL write misses  \\ \midrule
no-heuristic ($d=8$) & 122'150'286 & 14'777'070 \\
greedyheuristic ($d=8$) & 69'653'838 & 12'328'994 \\
no-heuristic ($d=256$) & 450'209'609 & 20'438'131\\
\bottomrule
\end{tabular}
\caption{Cachegrind results for versions of our implementation with (\textit{greedyheuristic}) or without (\textit{no-heuristic}) memory reordering on the \textit{Synthetic Clustered Dataset} ($n=131'072$ and $16$ clusters) for the specified dimension. Increasing $d$ by a factor of 32 increases the last-level read misses by a smaller factor.}
\label{table:cachegrind}
\end{table}

%So we can directly see a major improvement of locality with our greedy clustering heuristic by comparing the resulting misses of the first and second line in the table \ref{table:cachegrind}.
A major improvement of locality when using the greedy clustering heuristic can be seen in terms of the cache misses. Our heuristic nearly halves the number of LL cache data read misses. The increase in operational intensity by the reduction in cache misses moves the computation to the right in the roofline plot (\emph{no-heuristic, dim8} to \emph{greedyheuristic, dim8} in figure \ref{fig:roofline}). 

If we increase the dimension we do more work $W(n)$ and achieve a higher operational intensity because the last-level read misses increase by a smaller factor. Since features for a single datapoint are continuous in memory, we attribute this to spatial locality.
%Remember that we always load a full cache block for a miss which are 64 bytes = 16 floats, which is continuous in memory for our higher dimensional points and therefore the impact on the data movement is not that big compared to the higher work as you can see by comparing the misses of the first line to the last line in table \ref{table:cachegrind}.

%On a cache miss an entire cache block containing 16 single precision floats which are continuous in memory are loaded. For higher dimensional data points the impact on the data movement is small compared to ???. This is visible....

\subsection{Greedy Clustering Heuristic Evaluation}
\begin{figure}\centering
  \includegraphics[width=0.4\textwidth]{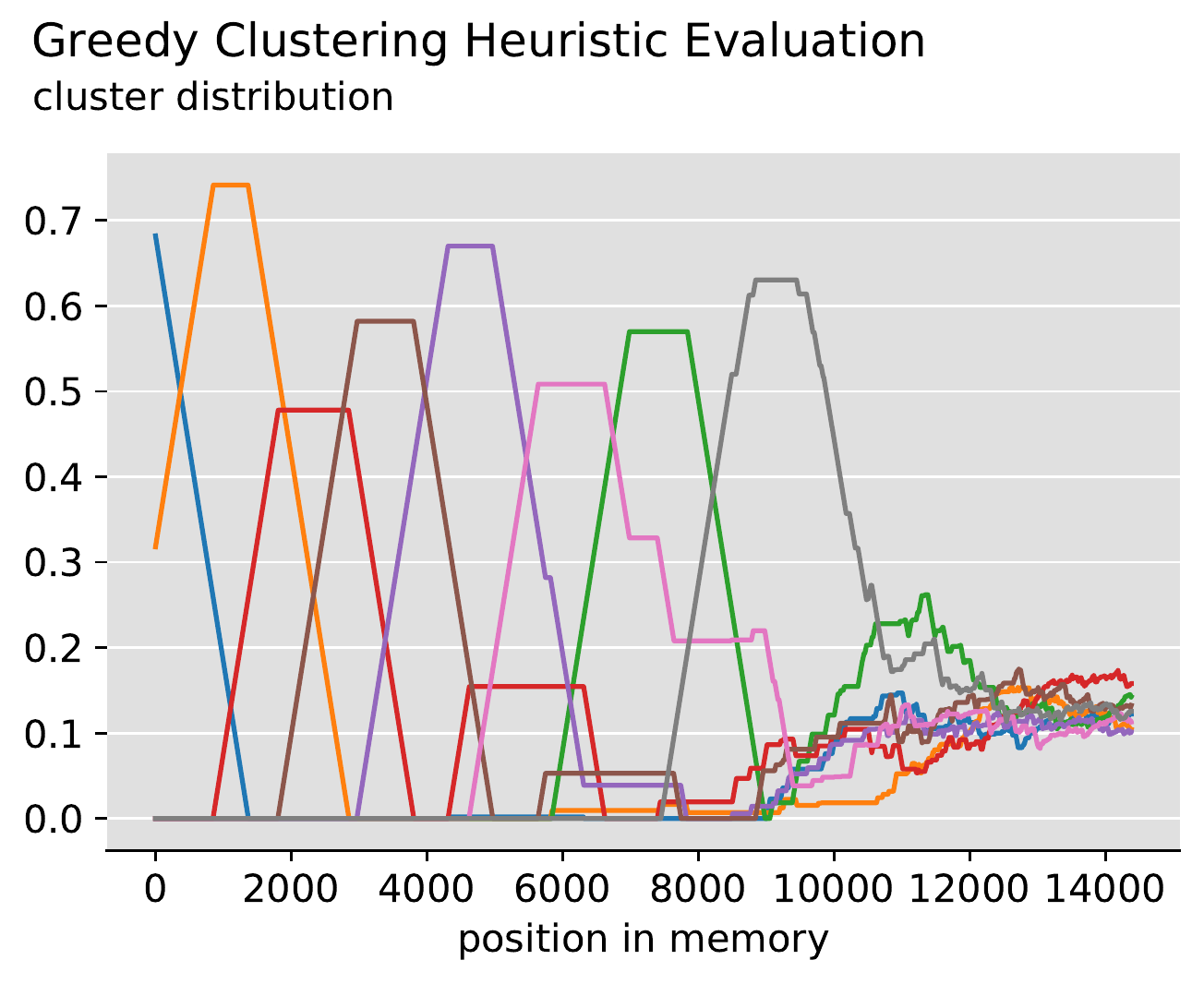}
  \caption{Each line represents the fraction of datapoints belonging to single cluster in a 2000 sized window (y-axis) starting at the position given in the x-axis. Plot represents the cluster distribution \textit{greedyclustering} reordering retrieved on our \textit{Synthetic Clustered Dataset} with parameters $n=16'384$, $d=8$ and 8 clusters.\label{fig:clustering}}
\end{figure}

\mypar{Quality of Clustering with Greedy Heuristic} 
%The whole discussion of the greedy clustering algorithm begs the question whether we actually recover the clusters or just achieve a speedup due to other reasons. 
This section investigates both how well the greedy clustering algorithm recovers the clusters of the underlying dataset and how this causes the speedup.

To test this, we ran our greedy clustering algorithm on the \textit{Synthetic Clustered Dataset}. We then reordered our data according to the permutation $\sigma$ found by the greedy clustering algorithm. 
%We can see in \ref{fig:clustering} that our algorithm recovers many of the clusters in the beginning of the dataset, where it still has many unassigned nodes left. 
Figure \ref{fig:clustering} shows that the algorithm successfully recovers many of the clusters in the beginning of the dataset, where the heuristic can chose from many unassigned nodes.

Towards the end of the dataset all of the relative frequencies of the eight clusters are around $\frac{1}{8}$ - our heuristic stops working. This is expected since our algorithm is restricted by design to a single pass through the data. If a certain cluster was already handled and most of the points belonging to it were moved to some location, then the missing few will end up at the end of our simplified memory layout.

%\begin{figure}\centering[h!]
\begin{figure}[h!]
  \includegraphics[width=0.4\textwidth]{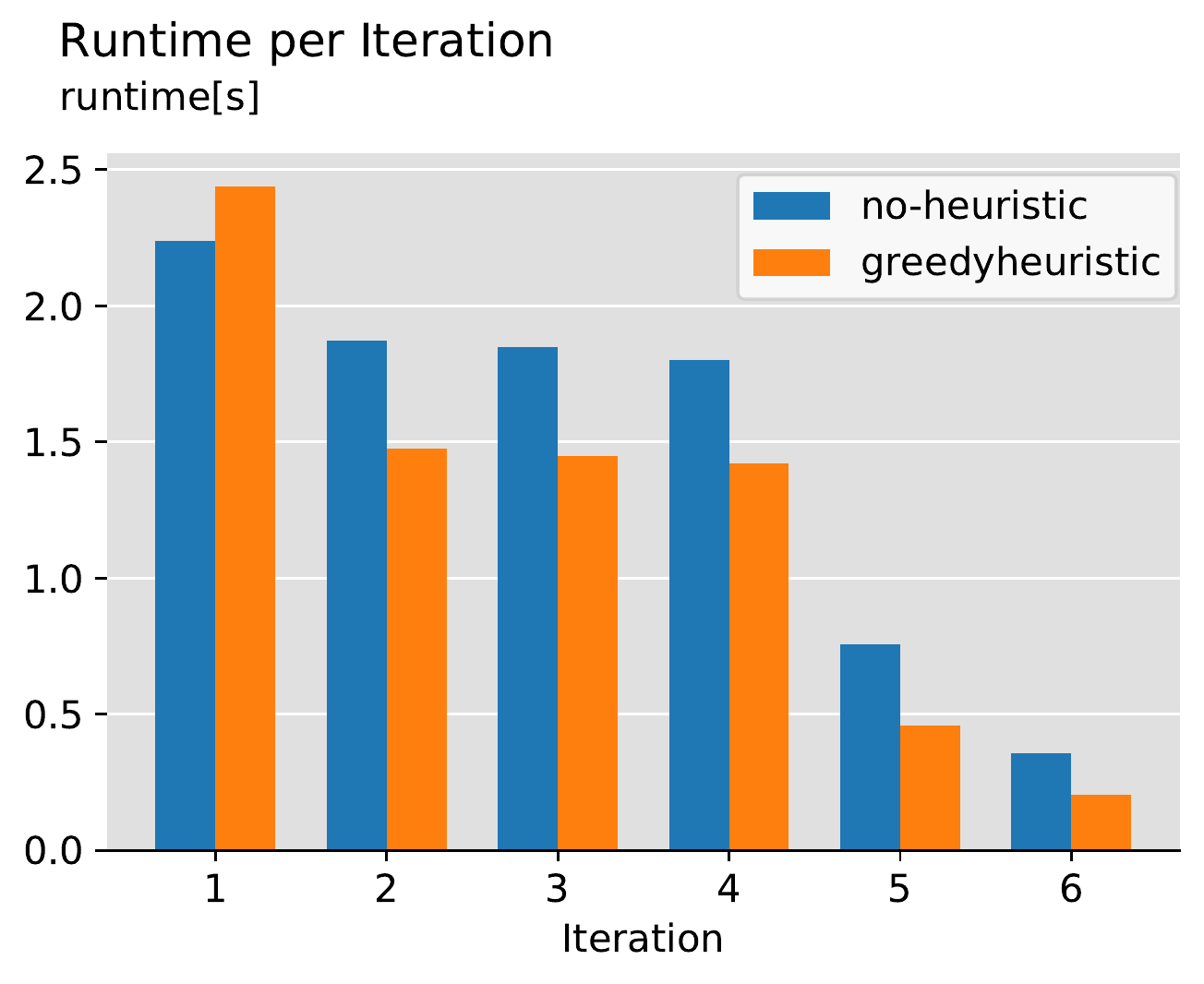}
  \caption{Time spent on each iteration on {Synthetic Clustered Dataset} with 16'384 points, 16 clusters and 8 dimensions. While the first iteration is slower due to the overhead of the heuristic, we profit in later iterations.\label{plot:iterations}}
\end{figure}

\mypar{Empirical Evaluation on Synthetic Dataset}
To test whether our heuristic leads to a speedup when the clustered assumption is satisfied we created such a dataset (\textit{Synthetic Clustered Dataset}).
%In order to put our assumption to the test we constructed a datset which satisfies the clustered assumption (\textit{Synthetic Clustered Dataset}).
We ran our algorithm twice on this dataset, once with greedy clustering and subsequent reordering of the data in memory and once without those changes. As reasoned in the previous section, we expect a shorter running time when doing the reordering. One can see in Figure \ref{plot:iterations} that the first iteration takes slightly longer to finish in the greedy clustering version of our algorithm due to overhead, but we can profit from the reordering in all subsequent iterations where we are consistently faster than the non reordered version. In total this amounts to a speedup of 18.46\% over all iterations.

\begin{table}\centering
\begin{tabular}{@{}ccc@{}}\toprule
runtime & MNIST & Audio \\ \midrule
no-heuristic & 12.12s & 4.78s\\
greedyheuristic & 11.45s & 4.53s\\
PyNNDescent & 24.41s & 14.47s\\
\bottomrule
\end{tabular}
\caption{Runtimes on the real-world \textit{MNIST} and \textit{Audio} datasets. Note how even though the clustered assumption might not hold, we still get a speedup on both datasets using our greedy clustering and subsequent reordering of data in memory. Our final implementation \textit{greedyclustering} is significantly faster on both datasets than PyNNDescent.\label{tbl:clusteredrealworld}}
\end{table}

\mypar{Real World Data Evaluation}
As previously hinted at we observed a speedup on real world data when applying our clustering heuristic. Consult Table \ref{tbl:clusteredrealworld} to see the runtimes on the \textit{MNIST} and \textit{Audio} datasets. According to our observations, the clustering heuristic does not manage to cluster MNIST semantically. Still, the reordering does improve locality which makes sense considering we move nodes together which are close in data space.

\subsection{Performance}
\begin{figure}\centering
  \includegraphics[width=0.4\textwidth]{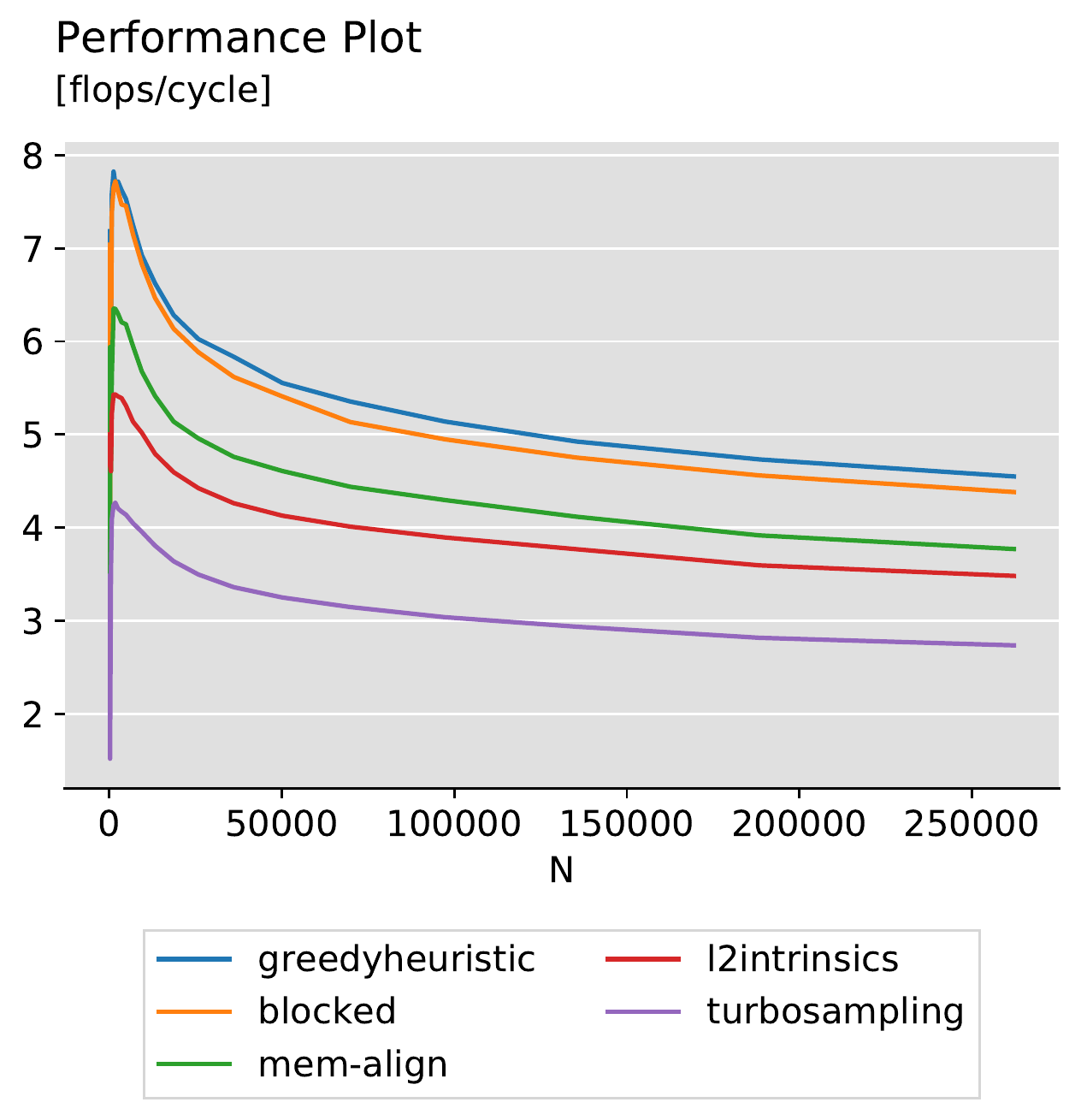}
  \caption{Performance of the discussed improvements on the \textit{Synthetic Gaussian Dataset}. The dimensions is fixed at $256$, while the $n$ increases along the x-axis. Lines correspond to specific tags of our code where every versions contains the improvements of previous versions to showcase the effect of specific changes.\label{plot:performanceplot256}}
\end{figure}

To show how the performance scales with number of input points $n$, we used the \textit{Synthetic Gaussian Dataset} with dimension $256$. Every line in Figure \ref{plot:performanceplot256} corresponds to a specific version of our code, building on top of the improvements of the lines below.
\\
\\
\textbf{turbosampling} Selection step improvements (Section \ref{subsec:selection_step})\\
\textbf{l2intrinsics} Optimized L2 distance calculation using intrinsics (Section \ref{subsec:highdimension})\\
\textbf{mem-align} Data aligned to 256bits (Section \ref{subsec:highdimension})\\
\textbf{blocked} Blocking of the L2 distance calculations (Section \ref{subsec:highdimension})\\
\textbf{greedyheuristic} Reordering of memory using heuristic (Section \ref{subsec:greedy})\\

%Performance degrades rapidly at first, this is expected since we outgrow our caches pretty quickly. 
Even for small $n$ the performance rapidly degrades as cache sizes are exceeded.
Note how all of the improvements lead to speedups, resulting in a performance gain of 1.5x. Considering the difficult setting of a randomized approximation algorithm, this is a remarkable improvement from our \textit{turbosampling} baseline.

\subsection{Impact of Dimensionality}
\begin{figure}\centering
  \includegraphics[width=0.4\textwidth]{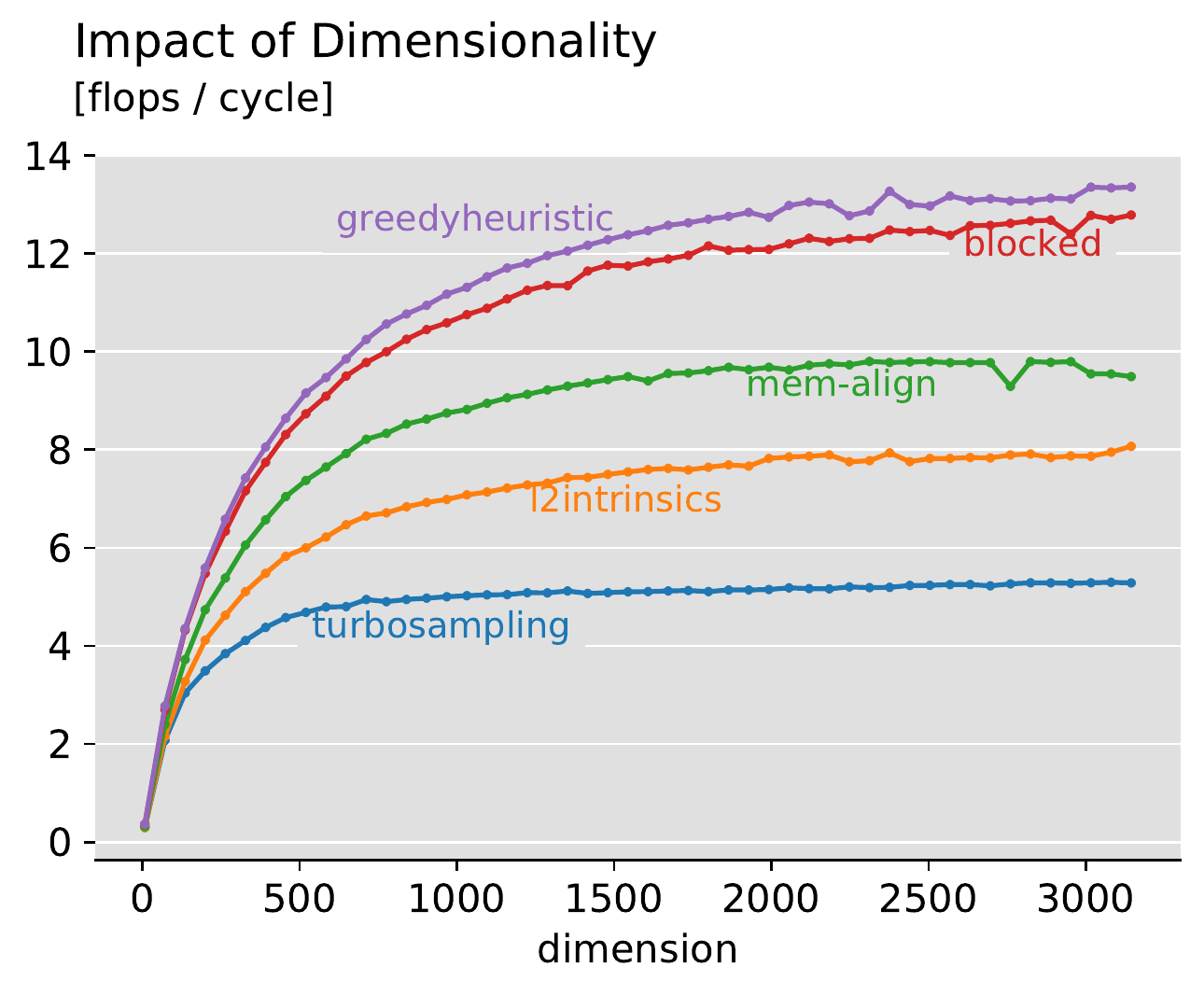}
  \caption{Performance for the various versions of our code on the \textit{Synthetic Single Gaussian Dataset} with $n=16'384$ and increasing dimension along the x-axis.\label{plot:dimensionality}}
\end{figure}

We previously argued that increasing the dimension of our input data makes our implementation compute bound. In low dimensions, our algorithm spends a large portion of its runtime in the selection step. For higher dimensions, the focus changes to the calculation step. Some improvements can therefore only unfold their true potential when we increase the dataset dimensionality. In Figure \ref{plot:dimensionality} we kept the number of datapoints $n$ fixed at $16'384$ and varied the dimension from $8$ up to $3144$ in increments of $64$. We used our \textit{Synthetic Single Gaussian Dataset} for this evaluation. Note how \textit{turbosampling} only sees a 3.52x performance gain when increasing the dimension while our optimizations targeted at high-dimensional data profit much more from the higher dimension - \textit{blocked} sees a 8.90x speedup going from $d=8$ to $d=3144$. Those results indicate that the impact of our optimizations varies with the dimension as expected.

\section{Conclusion}
NN-Descent is an efficient algorithm for computing K-NNGs.
The presented fast single-core C implementation incorporates numerous orthogonal optimizations which improve the runtime significantly in low-dimensional, as well as high dimensional usecases.\\
The presented optimizations to the selection step which reduce the number of memory passes and simplify the datastructures are particularly primarily for low dimensional data.\\
For high dimensional input data the implementation becomes compute bound and beneficial memory alignment and the use of blocked distance evaluations become paramount. Blocking is made possible by the restriction to the l2-distance while restricing the dimensionality makes memory alignment cheaper.\\
An interesting topic for future work would be to further explore heuristics for reordering the data. The evaluation shows that such heuristics can improve performance even on real world data. Such reordering heuristics are made possible by the iterative nature of the NN-Descent algorithm.\\
The significantly lower runtime on real world datasets compared to the popular PyNNDescent implementation illustrates the value of specialized implementations for common usecases.

% BiBTeX files (here: bibl_conf). The IEEEbib.bst bibliography
%\bibliographystyle{IEEEbib}
%\bibliographystyle{IEEE}
%\bibliography{bibl_conf}
\printbibliography

\end{document}